\documentclass[10pt,twocolumn,letterpaper]{article}
\usepackage{cvpr}
\usepackage{times}
\usepackage{epsfig}
\usepackage{graphicx}
\usepackage{amsmath}
\usepackage{amssymb}
\usepackage{subfig}
\usepackage{enumitem}
\usepackage{authblk}

\usepackage[breaklinks=true,bookmarks=false]{hyperref}

\cvprfinalcopy 

\def\cvprPaperID{****} 

\begin{document}

\title{Domain-Invariant Adversarial Learning for Unsupervised Domain Adaption}

\author[1]{Yexun Zhang}
\author[1]{Ya Zhang}
\author[1]{Yanfeng Wang}
\author[2]{Qi Tian}
\affil[1]{Cooperative Medianet Innovation Center,  Shanghai Jiao Tong University}
\affil[2]{Huawei Noah's Ark Lab}
\affil[ ]{\tt\small\{zhyxun,ya$\_$zhang,wangyanfeng\}@sjtu.edu.cn, tian.qi1@huawei.com}

\maketitle

\begin{abstract}
Unsupervised domain adaption aims to learn a powerful classifier for the target domain given a labeled source data set and an unlabeled target data set. To alleviate the effect of `domain shift', the major challenge in domain adaptation, studies have attempted to align the distributions of the two domains.
Recent research has suggested that generative adversarial network (GAN) has the capability of implicitly capturing data distribution.
In this paper, we thus propose a simple but effective model for unsupervised domain adaption leveraging adversarial learning. The same encoder is shared between the source and target domains which is expected to extract domain-invariant representations with the help of an adversarial discriminator. With the labeled source data, we introduce the center loss to increase the discriminative power of feature learned. We further align the conditional distribution of the two domains to enforce the discrimination of the features in the target domain. Unlike previous studies where the source features are extracted with a fixed pre-trained encoder, our method jointly learns feature representations of two domains.
Moreover, by sharing the encoder, the model does not need to know the source of images during testing and hence is more widely applicable. We evaluate the proposed method on several unsupervised domain adaption benchmarks and achieve superior or comparable performance to state-of-the-art results.
\end{abstract}

\section{Introduction}

Deep neural networks have drawn broad attention due to its impressive performance on a variety of tasks. Training a deep neural network usually requires a large labeled dataset. However, collecting and annotating a dataset for each new task is time-consuming and expensive. Fortunately, there are often a large amount of data available from other related domains and tasks and using the auxiliary data may alleviate the necessity of annotating a new dataset. However, due to factors such as image condition and illumination, datasets from two domains usually have different distributions. When the model trained on one dataset is tested on the other, the performance often greatly drops due to the `domain shift' problem. Domain adaption, as a sub-line of transfer learning, aims to solve the `domain shift' problem.

For unsupervised domain adaption, where all samples in the target domain are unlabeled, many studies try to align the statistical distributions of the source and target domains using various mechanisms, such as maximum mean discrepancy (MMD)~\cite{long2015learning,long2016unsupervised,pan2011domain}, correlation alignment (CORAL)~\cite{sun2016return,sun2016deep} and Kullback-Leibler (KL) divergence~\cite{zhuang2015supervised}.
Recently, adversarial learning is adopted to align the distributions by extracting features which are indistinguishable by the domain discriminator \cite{tzeng2017adversarial}. Usually  two separate encoders are trained, one for source domain and one for target domain.
The source encoder is usually pretrained first and fixed during domain adaption.

\begin{figure*}[!thp]
    \centering
    \includegraphics[height=1.9in,width=6in]{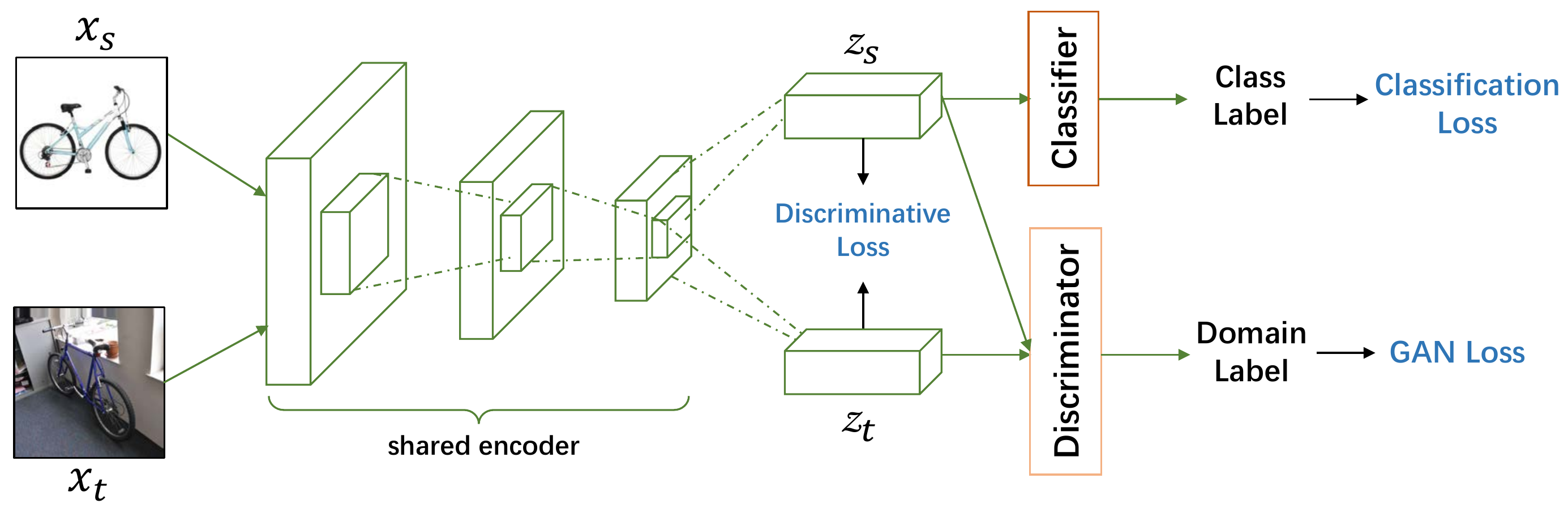}
    \caption{The detailed architecture of the proposed DIAL network for unsupervised domain adaption.}
    \label{fig:network}
    \vspace{-10pt}
\end{figure*}

In this paper, we propose a simple but effective model for unsupervised domain adaption. Inspired by the fact that humans can correctly recognize an object without being aware of its domain, we design a Domain-Invariant Adversarial Learning (DIAL) network, consisting of an encoder, a classifier, and a discriminator (Fig. \ref{fig:network}), for representations that are both domain-invariant and discriminative. Unlike the models using two separate feature extractors for the source and target domains, DIAL shares a single encoder between two domains and has no need to know the source of images during testing. The extracted features are then sent to the adversarial discriminator. The encoder and the discriminator play a min-max game, with the goal that the source of features cannot be distinguished by the discriminator. In this way, the encoder is expected to learn domain-invariant representations and ignore domain-specific information.
Furthermore, to enforce the discriminative power of feature representations, with the labeled data in the source domain, we introduce the center loss.
We also align the conditional distribution of the source and target domains, which has been largely ignored by existing adversarial-based domain adaption methods. However, aligning conditional distribution $P(Y|X)$ is quite challenging due to the absence of labels in target domain. We thus resort to the pesudo labels in target domain and align the class-conditional distribution $P(X|\hat{Y})$, which is expected to guide the target features to fall into correct clusters.
With the above design, the feature representations of the two domains are learned simultaneously, unlike previous studies where the source features are fixed during adaptation.

We evaluate the proposed DIAL on several unsupervised benchmarks and achieve new state-of-the-art results.
The main contributions of this paper are summarized as following.
\begin{itemize}[nosep, wide=0pt, leftmargin=*, after=\strut]
\item We propose a simple but effective model for unsupervised domain adaption, which is shown to extract domain-invariant and discriminative features for source and target images.
\item The model shares one feature extractor between two domains and has no need to know the source of images during testing.
\item We introduce the center loss in order to learn more discriminative feature representations.
\item Besides the marginal distribution, we also align the conditional distributions of source and target domain.
\item We evaluate the proposed method on several unsupervised domain adaption benchmarks and achieve new state-of-the-art results.
\end{itemize}
\section{Related Work}
For unsupervised domain adaption, the main approach is to guide the feature learning by minimizing the difference between the distributions of source domain and target domain. Several methods have used the MMD to measure the difference of distributions. \cite{pan2011domain} proposed the transfer component
analysis (TCA) to minimize the discrepancy of two domains in a Reproducing Kernel Hilbert Space
(RKHS) using MMD. Then, \cite{long2015learning,tzeng2014deep} extended the MMD to deep neural networks and achieved great success.
Rather than using a single adaption layer and linear MMD, Long et al. \cite{long2015learning} proposed
the deep adaptation network (DAN) which matches the shift in marginal distributions across domains by adding multiple adaptation layers and exploring multiple kernels. Further, Long et al. proposed a joint adaptation network (JAN) \cite{long2016deep} which aligns the shift in the joint distributions of input images and output labels.
Different than MMD, CORAL \cite{sun2016return} learns a linear transformation that aligns convariance of the source and target domains. Then, Sun et al. \cite{sun2016deep} extended CORAL to deep neural networks. Another commonly used metric to measure the discrepancy between domains is central moment discrepancy (CMD) \cite{zellinger2017central}, which restrains the domain discrepancy by matching the higher-order moments of the domain distributions.

Inspired by the generative adversarial networks (GANs) \cite{goodfellow2014generative}, adversarial learning is introduced to restrain the domain discrepancy by learning representations which is simultaneously discriminative in source labels and indistinguishable in domains. Tzeng et al. proposed the adversarial discriminative domain adaption (ADDA) \cite{tzeng2017adversarial}, which uses GANs to train an encoder for target samples, by making the features extracted with this encoder indistinguishable from the ones extracted through an encoder trained with source samples. Then, Volpi et al. \cite{volpi2017adversarial} extended the ADDA framework by forcing the learned feature extractor to be domain-invariant and training it through data augmentation in the feature space.

Other methods have chosen generative methods to minimize the domain discrepancy. In \cite{ghifary2016deep}, a deep reconstruction-classification network (DRCN) is introduced to learn common representations for both domains through the joint optimization of supervised classification of labeled source data and unsupervised reconstruction of unlabeled
target data. In \cite{bousmalis2016domain}, Bousmalis et al. proposed the domain separation network (DSN) which explicitly
learns to extract image representations that are partitioned into two components, one for the private information of each domain (domain feature) and the other for the shared representation across domains (content feature), to reconstruct the images and features from both domains.

Recently, image-to-image translation based methods are proposed for domain adaption by transferring images into target domain and then directly training classifiers on them. Taigman et al. proposed the Domain Transfer Network (DTN) \cite{taigman2016unsupervised} which is optimized by a compound loss function including a multi-class GAN loss, an f-constancy component, and a regularizing component that encourages the transfer network to map samples from target domain to themselves. This network can transfer one image from the source domain to the target domain. In \cite{liu2016coupled}, Liu and Tuzel introduced the coupled GANs (CoGAN) which can learn a joint distribution across multiple domains without requirement for paired images.
CoGAN consists of a pair of GANs and each has a generative model for synthesizing realistic images in one domain and a discriminative model for classifying whether an image is real or synthesized.
It can be applied for domain adaption by attaching a softmax layer to the last hidden layer of the discriminator,
and jointly solving the classification problem in the source domain and the CoGAN learning problem.
As an extension of CoGAN, Liu et al. proposed the unsupervised image-to-image translation (UNIT) \cite{liu2017unsupervised} network, which combined the GANs with variational auto-encoders (VAEs) and achieved unsupervised image-to-image translation based on the shared-latent space assumption. Lu et al. proposed the duplex GAN (DupGAN) \cite{hu2018duplex} to achieve domain invariant feature extraction and domain transformation. Cycle-Consistent Adversarial Domain Adaptation (CyCADA) proposed in \cite{hoffman2017cycada} adapts representations at both the pixel-level and feature-level while enforcing semantic consistency, which achieved satisfying performance on both digital classification and semantic segmentation.

\section{Model}
In this section, we introduce the proposed DIAL network in detail, whose architecture is displayed in Fig.~\ref{fig:network}. The whole network is elegant and consists of an encoder, a classifier and an adversarial discriminator. Our goal is to learn both domain-invariant and discriminative features which will benefit the domain adaption. In the following, we will introduce how we achieve this.

\subsection{Domain-invariant Feature Extraction}
For unsupervised domain adaption, extracting domain-invariant features is critical to alleviate the effect of the domain shift. In previous studies \cite{rozantsev2018beyond,tzeng2017adversarial}, totally separate or partially tied feature extractors are usually used for source domain and target domain. However, in this study, we find that sharing a single feature extractor between source and target domain seems more effective to learn domain-invariant features for unsupervised domain adaption.

In Figure~\ref{fig:network}, the images from the source domain and target domain are passed through one shared encoder $E$ and we aim to extract features which only contain the information about the content of the image, namely domain-invariant features. Here, we adopt the adversarial learning and add a discriminator to distinguish which domain the extracted feature is from and simultaneously, the encoder tries to extract features which are indistinguishable for the discriminator. The following adversarial loss is applied:
\begin{align}
    \min_{\theta_E} \max_{\theta_D} \mathcal{L}_{GAN} & = \sum_{x_i\in X_s}logD(E(x_i)) \\ \nonumber
    & + \sum_{x_i\in X_t}log(1-D(E(x_i)))
\end{align}
where $D(\cdot)$ is the probability of being source features predicted by the discriminator $D$, $\theta_E$ and $\theta_D$ are parameters of encoder $E$ and discriminator $D$, $X_s$ and $X_t$ are distributions of samples in source domain and target domain, respectively.

Although the inputs of the encoder are from two different domains, the extracted features cannot be distinguished by the discriminator about domains. By this limited condition, we expect the encoder to only extract the content information which is shared between these two domains and ignore the private domain information. Besides, with the shared encoder, the model can receive images from both source domain and target domain and we do not need to know the source of images during testing.

\subsection{Discriminative Feature Extraction}

Since we have labels for samples in the source domain, the features of the source domain will be classified by the classifier $C$, which is a fully connected softmax layer with the size dependent on the task. The optimization function for the classification of the labeled data in source domain is defined as:
\begin{equation}
    \min_{\theta_E,\theta_C} \mathcal{L}_{s} = \sum_{(x_i,y_i)\in (X_s,Y_s)} H(C(E(x_i)),y_i)
\end{equation}
where $H(\cdot)$ is the cross entropy loss used in the softmax layer, ($X_s$, $Y_s$) is the distribution of samples and labels in the source domain and $\theta_C$ are parameters of the classifier.

Furthermore, it is important to keep the discriminative power of feature representations during domain adaption. Although the distributions of source and target domains are aligned, there may sill be some samples falling into inter-class gaps, which proposes the requirement for learning more discriminative features. In the learning literature, there exists several methods for learning discriminative features, such as the triplet loss \cite{schroff2015facenet}, the contrastive loss \cite{sun2014deep} and the center loss \cite{wen2016discriminative}. Both the triplet loss and the contrastive loss need to construct a lot of image pairs and compute the distance between images of each pair, which is computationally complicated. Therefore, in this study, we introduce the center loss, which can be flexibly combined with the above classification loss.

For samples in source domain which have labels, we adopt the following loss to cluster the features belonging to the same class:
\begin{equation}
   \min_{\theta_E} \mathcal{L}_{cs} = \sum_{(x_i, y_i)\in (X_s, Y_s)} ||E(x_i) - c_{y_i} ||_2^2
\end{equation}
where $c_{y_i}$ is a $d$-dimensional vector representing the center of the $y_i$-th class. Ideally, each class center should be calculated using the features of all samples belonging to that class. But due to we optimize the model with mini-batch samples, it is difficult to compute the average of all samples. Therefore, we first initialize the class center by the batch in the first iteration, then update the centers by the following strategy:
\begin{equation}
    c_k^{t+1} = c_k^t - \gamma \Delta c_k^t, \quad k=1,2,\ldots,K
\end{equation}
where $c_k^t$ is the center for the $k$-th class in iteration $t$, $\gamma$ is the learning rate for updating the centers, $K$ is the total number of classes and
\begin{equation}
    \Delta c_k^t =  \frac{\sum_{(x_i,y_i)\in \mathcal{B}^t} \mathbb{I}(y_i=k) (c_k^t-E(x_i))}{1+N_k}
\end{equation}
where $\mathcal{B}^t$ represents the mini-batch in iteration $t$, $\mathbb{I}(\cdot)$ is an indicator function and $N_k=\sum_{(x_i,y_i)\in \mathcal{B}^t}$ $\mathbb{I}(y_i=k)$ is the number of samples in batch $\mathcal{B}^t$ which belong to class $k$.

\subsection{Conditional Distribution Alignment}
In existing adversarial-based domain adaption methods, only marginal distribution adaption is concerned by aligning the distribution $P(X)$. However, as verified in some previous research \cite{long2013transfer}, the conditional distribution $P(Y|X)$ of two domains may also be different. Since we have no labels for target domain, directly aligning the $P(Y|X)$ is challenging. Inspired by \cite{long2013transfer,volpi2017adversarial}, we explore the pseudo labels of target samples and resort to explore the sufficient statistics of class-conditional distributions $P(X|Y)$ instead as done in \cite{long2013transfer}.

Therefore, for unlabeled samples in target domain, we assign each sample with the pesudo label predicted by the source classifier and define the following loss:
\begin{equation}
   \min_{\theta_E} \mathcal{L}_{ct} = \sum_{x_i\in \Phi(X_t)} ||E(x_i) - c_{\hat{y}_i}||_2^2
\label{eq:mag}
\end{equation}
where $\hat{y}_i$ is the label of $x_i$ predicted by the classifier $C$. Since not all the predicted labels are accurate, we calculate the above $\mathcal{L}_{ct}$ only on $\Phi(X_t)$ which is a subset of $X_t$ and the samples in it satisfy:
\begin{equation}
    \Phi(X_t) = \{ x_i| x_i\in X_t \; \textrm{and} \; \max(p(x_i)) \geq T\}
\end{equation}
where $p(x_i)$ is a $K$-dimensional vector with the $i$-th dimension being the predicted probability of belonging to $i$-th class, $\max(p(x_i))$ is the probability of sample $x_i$ belonging to the predicted class, and $T$ is the threshold we set. By this way, samples in target domain are expected to fall into corresponding clusters.

Therefore, the total objective function of the model can be formulated as:
\begin{equation}
    \min_{\theta_E, \theta_C} \max_{\theta_D} \mathcal{L}_{GAN} + \alpha \mathcal{L}_s  + \beta_1 \mathcal{L}_{cs} + \beta_2 \mathcal{L}_{ct}
\end{equation}
where $\alpha$, $\beta_1$ and $\beta_2$ are weighted parameters.


\begin{table*}[!htpb]
    \centering
    \setlength{\tabcolsep}{1.2mm}
    \setlength{\abovecaptionskip}{-1pt}
    \begin{tabular}{c|ccccccc}
        \hline
        Method &  MNIST$\rightarrow$USPS(P1) & USPS$\rightarrow$MNIST(P1) & MNIST$\rightarrow$USPS(P2) & USPS$\rightarrow$MNIST(P2) & SVHN$\rightarrow$MNIST \\
        \hline
        Source &  74.59$\pm$1.30 & 60.54$\pm$1.33 & 86.33$\pm$0.47 & 67.49$\pm$1.34  & 67.95$\pm$0.89 \\
        \hline
        DANN \cite{ganin2014unsupervised}    &    77.1$\pm$1.8 & 73.0$\pm$2.0 & -  &  - & 73.9 \\
        ADDA \cite{tzeng2017adversarial}   &    89.4$\pm$0.2 & 90.1$\pm$0.8 & -  &  - & 76.0$\pm$1.8 \\
        UNIT \cite{liu2017unsupervised}    &    -            &              & 95.97 &  93.58  & 90.53 \\
        CoGAN \cite{liu2016coupled} &    91.2$\pm$0.8 & 89.1$\pm$0.8 & 95.65 &  93.15  & - \\
        DI \cite{volpi2017adversarial} & 91.4$\pm$0.0 & 87.9$\pm$0.5 & 95.4$\pm$0.2 & - & 85.1$\pm$2.6 \\
        DIFA  \cite{volpi2017adversarial}  &    92.3$\pm$0.1 & 89.7$\pm$0.5 & 96.2$\pm$0.2 & - & 89.2$\pm$2.0 \\
        CyCADA \cite{hoffman2017cycada} &    -        & -        &  95.6$\pm$0.2 & 96.5$\pm$0.1 & 90.4$\pm$0.4 \\
        DupGAN \cite{hu2018duplex}  &    -        & -        & 96.0  & 98.75 & 92.46 \\
        SimNet \cite{pinheiro2018unsupervised} & - & - & 96.4 & 95.6 & - \\
        ACGAN \cite{sankaranarayanangenerate} & 92.8 & 90.8 & 95.3 & - & 92.4 \\
        \hline
        DIAL (ours) & 95.02$\pm$0.22 & 97.28$\pm$0.25 & 97.06$\pm$0.20 & 99.12$\pm$0.06 & 95.85$\pm$0.81 \\
        \hline
        Target &  95.65$\pm$0.14 & 97.83$\pm$0.49 & 96.71$\pm$0.18 & 99.36$\pm$0.04 & 99.36$\pm$0.04 \\
        \hline
    \end{tabular}
    \vspace{5pt}
    \caption{Comparison results of different models for unsupervised domain adaption on the digital dataset.}
    \label{tab:digital}
    \vspace{-10pt}
\end{table*}

\section{Experiments}
In this section, we evaluate the proposed method by comparing it with several state-of-the-art methods for unsupervised domain adaption. We first introduce the datasets we used. Then, we introduce the implementation details and finally, the experimental results are analyzed in detail.

\subsection{Datasets}
\textbf{Digital Dataset} For digit classification, the datasets of MNIST \cite{lecun1998gradient}, USPS \cite{denker1989neural} and SVHN \cite{netzer2011reading} are used for evaluating all of the methods. All three datasets contain images of digits 0-9 but with different styles. MNIST is composed of 60,000 training
and 10,000 testing images. USPS consists of 7,291 training
and 2,007 testing images. SVHN contains 73,257 training,
26,032 testing and 531,131 extra training images.
Following \cite{volpi2017adversarial}, we evaluate the adaption between MNIST and USPS by setting two protocols. For the first protocol (P1), we follow the training protocol established in \cite{long2013transfer} and randomly sampled 2,000 MNIST images and 1,800 USPS images for training.
The second protocol (P2) uses the whole MNIST training set and USPS training set. For both P1 and P2, we evaluate the two directions of the split (MNIST$\rightarrow$USPS and USPS$\rightarrow$MNIST).
For the adaption of SVHN and MNIST, we only evaluate on SVHN$\rightarrow$MNIST following previous studies \cite{tzeng2017adversarial, volpi2017adversarial}.
All images in MNIST and USPS are transformed to be RGB with size 32$\times$32 which is the size of images in SVHN.

\textbf{Office-31 Dataset} The Office-31 dataset consists of 4,110 images spread across 31 classes in 3 domains: Amazon (2,817 images), Webcam (795 images), and Dslr (498 images). We compare the methods on all of the six combination pairs of the three domains: A$\leftrightarrow$W, A$\leftrightarrow$D and W$\leftrightarrow$D. Following previous work \cite{tzeng2017adversarial}, we train the model on all labeled data in source domain and all unlabeled data in target domain, and then test it on the data in target domain.

\textbf{ImageCLEF-DA Dataset} ImageCLEF-DA \cite{zhang2018collaborative} is a benchmark dataset for ImageCLEF 2014 domain adaptation challenge, which is collected by selecting the 12 common categories shared by the following three public datasets, each is considered as a domain: Caltech-256 (C), ImageNet ILSVRC 2012 (I), and Pascal VOC 2012 (P). There are 50 images in each category and 600 images in each domain. We follow \cite{long2016deep} and \cite{zhang2018collaborative} to report our results for six settings.

\subsection{Implementation Details}
For experiments on digital classification, we use the simple modified LeNet architecture following previous work \cite{tzeng2017adversarial}. The discriminator is composed of 3 fully connected layers: two layers with 500 hidden units followed by the final discriminator output. Each of the 500-unit layers uses a ReLU activation function.
The whole network is optimized by the RMSPropOptimizer with batch size 64 for MNIST$\leftrightarrow$USPS in P1 and 256 for other cases. We set the initial learning rate being 0.001 and decaying 0.5 times for very 60 epochs. The threshold $T$ is set to be 0.99 for all cases. We train the model progressively with $\alpha$=10, $\beta_1$=0.001 and $\beta_2$=0 for the first 30 epochs and then we change $\beta_1$=$\beta_2$=0.002 for another 30 epochs and then train the model with $\beta_1$=$\beta_2$=0.02 until convergence.

For Office-31 and ImageCLEF-DA dataset, we adopt the ResNet-50 \cite{he2016deep} as the base model which is pretrained on ImageNet, and the activations of the last layer pool5 are used as the image representations. Since the dataset is small, we only fine-tune the last block and the fully-connected layer of ResNet-50. We optimize the network by the Stochastic Gradient Descent (SGD) optimizer with the momentum of 0.9 and batch size 64, namely total 128 images from the source and target domain. We set the initial learning rate as 0.001 and decay 0.5 times for every 50 epochs to avoid over-fitting. Similarly, we first train the model with $\alpha$=10, $\beta_1$=0.001 and $\beta_2$=0 for 50 epochs and $\beta_1$=$\beta_2$=0.002 for another 50 epochs and then train the model with $\beta_1$=$\beta_2$=0.01 until convergence. The learning rate $\gamma$ for updating centers is set to be 0.5 in all cases. For fair comparison, we conduct each experiment for several times with random initialization and show the mean$\pm$std as the result.

\begin{table*}[!thb]
    \centering
    \setlength{\abovecaptionskip}{-1pt}
    \setlength{\tabcolsep}{3.6mm}
    \begin{tabular}{c|ccccccc}
        \hline
        Method &  A$\rightarrow$W & D$\rightarrow$W & W$\rightarrow$D & A$\rightarrow$D & D$\rightarrow$A & W$\rightarrow$A & Average \\
        \hline
        AlexNet \cite{krizhevsky2012imagenet} &   61.6$\pm$0.5 & 95.4$\pm$0.3 & 99.0$\pm$0.2 & 63.8$\pm$0.5 & 51.1$\pm$0.6 & 49.8$\pm$0.4 & 70.1 \\
        DDC  \cite{tzeng2014deep}  &    61.8$\pm$0.4 & 95.0$\pm$0.5 & 98.5$\pm$0.4 & 64.4$\pm$0.3 & 52.1$\pm$0.6 & 52.2$\pm$0.4 & 70.6 \\
        DAN \cite{long2015learning}   &    68.5$\pm$0.5 & 96.0$\pm$0.3 & 99.0$\pm$0.3 & 67.0$\pm$0.4 & 54.0$\pm$0.5 & 53.1$\pm$0.5 & 72.9 \\
        DANN \cite{ganin2014unsupervised}     & 73.0$\pm$0.5 & 96.4$\pm$0.3 & 99.2$\pm$0.3 & 72.3$\pm$0.3 & 53.4$\pm$0.4 & 51.2$\pm$0.5 & 74.3 \\
        JAN \cite{long2016deep} &    75.2$\pm$0.4 & 96.6$\pm$0.2 & 99.6$\pm$0.1 & 72.8$\pm$0.3 & 57.5$\pm$0.2 & 56.3$\pm$0.2 & 76.3 \\
        \hline
        VGG-16 \cite{simonyan2014very}  &    67.6$\pm$0.6 & 96.1$\pm$0.3 & 99.2$\pm$0.2 & 73.9$\pm$0.9 & 58.2$\pm$0.5 & 57.8$\pm$0.4 & 75.5 \\
        CMD \cite{zellinger2017central}  & 77.0$\pm$0.6 & 96.3$\pm$0.4 & 99.2$\pm$0.2 & 79.6$\pm$0.6 & 63.8$\pm$0.7 & 63.3$\pm$0.6 & 79.9 \\
        \hline
        ResNet-50 \cite{he2016deep} &    68.4$\pm$0.2 & 96.7$\pm$0.1 & 99.3$\pm$0.1 & 68.9$\pm$0.2 & 62.5$\pm$0.3 & 60.7$\pm$0.3 & 76.1 \\
        DDC \cite{tzeng2014deep}    &    75.6$\pm$0.2 & 96.0$\pm$0.2 & 98.2$\pm$0.1 & 76.5$\pm$0.3 & 62.2$\pm$0.4 & 61.5$\pm$0.5 & 78.3 \\
        ADDA \cite{tzeng2017adversarial}  &    75.1     & 97.0     & 99.6     & -        & -        & -        & -    \\
        DAN \cite{long2015learning}   &    80.5$\pm$0.4 & 97.1$\pm$0.2 & 99.6$\pm$0.1 & 78.6$\pm$0.2 & 63.6$\pm$0.3 & 62.8$\pm$0.2 & 80.4 \\
        DANN \cite{ganin2014unsupervised}  &    82.0$\pm$0.4 & 96.9$\pm$0.2 & 99.1$\pm$0.1 & 79.7$\pm$0.4 & 68.2$\pm$0.4 & 67.4$\pm$0.5 & 82.2 \\
        JAN \cite{long2016deep} &    86.0$\pm$0.4 & 96.7$\pm$0.3 & 99.7$\pm$0.1 & 85.1$\pm$0.4 & 69.2$\pm$0.4 & 70.7$\pm$0.5 & 84.6 \\
       SimNet \cite{pinheiro2018unsupervised}  & 88.6$\pm$0.5 & 98.2$\pm$0.2 & 99.7$\pm$0.2 & 85.3$\pm$0.3 & 73.4$\pm$0.8 & 71.8$\pm$0.6  & 86.2 \\
       ACGAN \cite{sankaranarayanangenerate}  & 89.5$\pm$0.5 & 97.9$\pm$0.3 & 99.8$\pm$0.4 & 87.7$\pm$0.5 & 72.8$\pm$0.3 & 71.4$\pm$0.4 & 86.5 \\
       iCAN \cite{zhang2018collaborative} & 92.5 & 98.8 & 100.0 & 90.1 & 72.1 &  69.9 & 87.2 \\
        \hline
        DIAL (ours) & 91.7$\pm$0.4 & 97.1$\pm$0.3  & 99.8$\pm$0.0 & 89.3$\pm$0.4 & 71.7$\pm$0.7 & 71.4$\pm$0.2 & 86.8\\
        DIAL (ours best) & 92.1 & 97.5 & 99.8 & 89.6 & 72.7 & 71.5 & 87.2 \\
        \hline
    \end{tabular}
    \vspace{5pt}
    \caption{Comparison results of different models for unsupervised domain adaption the Office-31 dataset.}
    \label{tab:office}
    \vspace{-10pt}
\end{table*}

\subsection{Results}
In this section, we will first display and analyze the experimental results on both digital classification and object recognition tasks. Then ablation study is conducted and finally, we visualize the features extracted by the encoder to further verify the proposed DIAL model.

\subsubsection{Results on Digital Classification}
For experiments on digital classification, we compare the proposed method with state-of-the-art methods to verify the effectiveness of the proposed method. The results are displayed in Table~\ref{tab:digital}. The row of ``Source" reports the accuracies on target data achieved by non-adapted classifiers trained on the source data. And the row of ``Target" reports the results on target data achieved by classifiers trained on target data. Since we follow the same experimental settings with most compared methods, we directly copy the results from corresponding papers.

As observed in Table~\ref{tab:digital}, deep transfer learning models perform better than non-adapted classifiers trained on the source data, indicating that integrating domain adaption modules into deep networks will help reduce the domain discrepancy. Among the deep transfer learning models, our proposed method outperforms all baselines on all tasks. In particular, our method improves the accuracy for a large margin even on difficult transfer tasks, e.g. SVHN$\rightarrow$MNIST, where the SVHN dataset contains significant
variations in scale, background, rotation and so on, and there is only slightly variation in the digits shapes, which makes it substantially different from MNIST dataset. These experimental results demonstrate that the proposed method is effective for unsupervised domain adaption with large scale datasets.
We owe the improvements to three points. First, we share one encoder between the source domain and target domain to extract domain-invariant representations and the features of source and target domains are jointly learned. Secondly, we extract discriminative features by integrating the classification loss with the center loss, which makes samples of the same class more compact. Thirdly, rather than only aligning the marginal distributions, we also align the class-conditional distribution $P(X|Y)$ of the two domains, by which samples in target domain will be guided to fall into corresponding class clusters.

\begin{table}[!t]
    \centering
    \setlength{\abovecaptionskip}{-1pt}
    \setlength{\tabcolsep}{0.8mm}
    \begin{tabular}{c|ccccccc}
        \hline
        Method &  I$\rightarrow$P & P$\rightarrow$I & I$\rightarrow$C & C$\rightarrow$I & C$\rightarrow$P & P$\rightarrow$C & Avg. \\
        \hline
        ResNet-50 \cite{he2016deep} & 74.6 & 82.9 & 91.2 & 79.8 & 66.8 & 86.9 & 80.4 \\
        DAN \cite{long2015learning} & 74.5 & 82.2 & 92.8 & 86.3 & 69.2 & 89.8 & 82.5 \\
        RTN \cite{long2016unsupervised} & 74.6 & 85.8 & 94.3 & 85.9 & 71.7 & 91.2 & 83.9 \\
        DANN \cite{ganin2014unsupervised}& 75.6 & 84.0 & 93.0 & 86.0 & 71.7 & 87.5 & 83.0 \\
        JAN \cite{long2016deep} & 76.8 & 88.0 & 94.7 & 89.7 & 74.2 & 91.7 & 85.8 \\
        iCAN \cite{zhang2018collaborative} & 79.5 & 89.7 & 94.7 & 89.9 & 78.5 & 92.0 & 87.4 \\
        \hline
        DIAL (ours) & 79.6 & 90.7  & 95.7 & 90.6 & 77.0 & 93.5 & 87.9 \\
        \hline
    \end{tabular}
    \vspace{5pt}
    \caption{Comparison results of different models for unsupervised domain adaption on the ImageCLEF-DA dataset. ``Avg." represents the average score.}
    \label{tab:ImageCLEF-DA}
    \vspace{-10pt}
\end{table}

\begin{table*}[!htpb]
    \centering
    \setlength{\tabcolsep}{0.3mm}
    \setlength{\abovecaptionskip}{-1pt}
    \begin{tabular}{c|ccccccc}
        \hline
        Method &  MNIST$\rightarrow$USPS(P1) & USPS$\rightarrow$MNIST(P1) & MNIST$\rightarrow$USPS(P2) & USPS$\rightarrow$MNIST(P2) & SVHN$\rightarrow$MNIST \\
        \hline
        $M_{\mathcal{L}_s+\mathcal{L}_{GAN}}$ & 91.56$\pm$0.91 & 93.81$\pm$1.36 & 96.29$\pm$0.39 & 98.89$\pm$0.11 & 88.34$\pm$0.89 \\
        $M_{\mathcal{L}_s+\mathcal{L}_{GAN}+\mathcal{L}_{cs}}$ & 93.52$\pm$0.81 & 96.89$\pm$0.99 & 96.06$\pm$0.30 & 99.06$\pm$0.05 & 89.27$\pm$1.76 \\
        $M_{\mathcal{L}_s+\mathcal{L}_{GAN}+\mathcal{L}_{cs}+\mathcal{L}_{ct}}$ & 95.02$\pm$0.22 & 97.28$\pm$0.25 & 97.06$\pm$0.20 & 99.12$\pm$0.06 & 95.85$\pm$0.81 \\
        \hline
    \end{tabular}
    \vspace{5pt}
    \caption{Ablation study on the digital dataset.}
    \label{tab:ablation1}
\end{table*}

\begin{table*}[!thb]
    \centering
    \setlength{\tabcolsep}{1.9mm}
    \setlength{\abovecaptionskip}{-1pt}
    \begin{tabular}{c|ccccccc}
        \hline
        Method &  A$\rightarrow$W & D$\rightarrow$W & W$\rightarrow$D & A$\rightarrow$D & D$\rightarrow$A & W$\rightarrow$A & Average \\
        \hline
        $M_{\mathcal{L}_s+\mathcal{L}_{GAN}}$ & 81.33$\pm$0.76 & 96.49$\pm$0.32 & 96.99$\pm$0.49 & 78.35$\pm$0.75 & 61.90$\pm$0.30 & 66.30$\pm$0.36 & 80.31\\
        $M_{\mathcal{L}_s+\mathcal{L}_{GAN}+\mathcal{L}_{cs}}$ &  90.00$\pm$0.96 & 96.18$\pm$0.23 & 99.68$\pm$0.1 & 84.48$\pm$1.18 & 68.64$\pm$0.82 & 69.08$\pm$0.96 & 84.68\\
        $M_{\mathcal{L}_s+\mathcal{L}_{GAN}+\mathcal{L}_{cs}+\mathcal{L}_{ct}}$ & 91.72$\pm$0.42 & 97.15$\pm$0.31  & 99.8$\pm$0.00 & 89.29$\pm$0.38 & 71.72$\pm$0.73 & 71.35$\pm$0.24 & 86.84\\
        \hline
    \end{tabular}
    \vspace{5pt}
    \caption{Ablation study on the Office-31 dataset.}
    \label{tab:ablation2}
    \vspace{-10pt}
\end{table*}

\subsubsection{Results on Office-31}
In contrast to digital classification which has a large dataset, object recognition on office-31 is a task where both source and target domains only have a small number of samples. The comparison results of the proposed method and state-of-the-art methods are shown in Table~\ref{tab:office}.
For fair comparison, the results of compared methods are copied from published papers.
The ``AlexNet", ``VGG-16" and ``ResNet50" display the results on target domain obtained by fine-tuning the AlexNet, VGG-16 and ResNet-50 respectively using the source data, which are pretrained on ImageNet. Besides the mean value and the standard deviation, we also show the best performance achieved by our model on each task in the row ``DIAL (ours best)".

As shown in Table~\ref{tab:office}, our proposed approach achieves superior or comparable performance to state-of-the-art results, demonstrating that our method is also effective for unsupervised domain adaption when source domain and target domain only have a small number of samples. It is worth noting that our method significantly improves the accuracies on difficult tasks compared with JAN, such as A$\rightarrow$W and A$\rightarrow$D where samples in source domain and target domain are very different, and W$\rightarrow$A and D$\rightarrow$A where the size of the source domain is very small. Besides, our method achieves comparable results with state-of-the-art methods, such as ACGAN and iCAN. However, our model is more elegant than ACGAN and iCAN, where ACGAN is based on the Auxiliary Classifier GAN and iCAN adopted complicated tricks to select pseudo-labelled target samples. In addition, models based on VGG-16 and ResNet-50 perform better than AlexNet-based models which implies that very deep models like VGG-16 and ResNet-50 not only learn better representations for general learning tasks but also learn more generalizable features for domain adaption.

\subsubsection{Results on ImageCLEF-DA}
Different from Office-31 where different domains are of different sizes, the three domains in ImageCLEF-DA are of equal size, which makes it a good complement to Office-31 for more controlled experiments.
The results on ImageCLEF-DA dataset are shown in Table~\ref{tab:ImageCLEF-DA}, where we only show the mean value of multiple experiments due to the limited space. As we can observe, the proposed method performs better than state-of-the-art methods on most cases except the C$\rightarrow$P, where iCAN performs best with the carefully selected pesudo-labelled target samples.

\subsubsection{Ablation Study}
To verify the effectiveness of each component in our model, we conduct ablation study on digital dataset and Office-31 dataset.
We trained another two models: one model is trained only using the softmax loss and the GAN loss which is denoted as ``$M_{\mathcal{L}_s+\mathcal{L}_{GAN}}$", and the other model is trained using the softmax loss, the GAN loss and the center loss for the source domain $\mathcal{L}_{cs}$, which is denoted as ``$M_{\mathcal{L}_s+\mathcal{L}_{GAN}+\mathcal{L}_{cs}}$". We compare the two models with the model trained with all losses and the results are shown in Table~\ref{tab:ablation1} and Table~\ref{tab:ablation2}.

As shown in these tables, after removing one or more parts, the performance degrades in most cases. The more parts are removed, the worse the performance is. This indicates that all parts are reasonably designed and they work harmoniously forming an effective solution for unsupervised domain adaption.
Comparing Table~\ref{tab:digital} and Table~\ref{tab:ablation1}, we can observe that ``$M_{\mathcal{L}_s+\mathcal{L}_{GAN}}$" performs better than models which use two different encoders for source domain and target domain, such as ADDA, indicating that sharing one feature extractor between two domains is better for unsupervised domain adaption to extract domain-invariant features.
In addition, ``$M_{\mathcal{L}_s+\mathcal{L}_{GAN}}$" performs better than DI model which also shares one encoder for source and target domains. The difference between the DI model and ``$M_{\mathcal{L}_s+\mathcal{L}_{GAN}}$" lies in the way of distribution alignment. In DI model, the features of the two domains extracted by the encoder will be aligned with features from a pretrained source encoder. But in ``$M_{\mathcal{L}_s+\mathcal{L}_{GAN}}$", the features sent into the discriminator are from the shared being trained encoder, namely the features of source domain and target domain are jointly learned. Experimental results demonstrate that jointly learning the feature representation for source domain and target domain is better.
In Table~\ref{tab:ablation1} and Table~\ref{tab:ablation2}, ``$M_{\mathcal{L}_s+\mathcal{L}_{GAN}+\mathcal{L}_{cs}}$" performs better than ``$M_{\mathcal{L}_s+\mathcal{L}_{GAN}}$" on most tasks, showing that keeping the discriminative power of learned representations in source domain helps to learn a better classifier.
Furthermore, after adding the loss $\mathcal{L}_{ct}$, the accuracies are improved further, indicating that aligning the class-conditional distribution $P(X|Y)$ is helpful for unsupervised domain adaption.

\begin{table}[!t]
    \centering
    \setlength{\tabcolsep}{1.0mm}
    \begin{tabular}{c|cc|cc}
        \hline
        Source &  $M_{\mathcal{L}_s}$ & $M_{\mathcal{L}_s+\mathcal{L}_{GAN}}$ &  $M_{\mathcal{L}_s+\mathcal{L}_{cs}}$ & $M_{\mathcal{L}_s+\mathcal{L}_{GAN}+\mathcal{L}_{cs}}$\\
        \hline
        M(P1) & 97.83 & 97.56 & 98.56 & 98.34\\
        M(P2) & 99.36 & 99.43 & 98.34 & 99.52\\
        U(P1)  & 95.65 & 95.28 & 96.51 & 96.62\\
        U(P2)  & 96.71 & 96.98 & 97.02 & 97.54\\
        S      & 92.44 & 92.18 & 93.34 & 91.16\\
        \hline
        Avg. & 96.40 & 96.29 & 96.75 & 96.64 \\
        \hline
    \end{tabular}
    \vspace{5pt}
    \caption{Results on test dataset in source domain classified with model $M_{\mathcal{L}_s}$ and $M_{\mathcal{L}_s+\mathcal{L}_{cs}}$ and the domain adapted models $M_{\mathcal{L}_s+\mathcal{L}_{GAN}}$ and $M_{\mathcal{L}_s+\mathcal{L}_{GAN}+\mathcal{L}_{cs}}$. M: MNIST, U: USPS, S: SVHN. ``Avg." represents the average accuracy.}
    \label{tab:com_source}
    \vspace{-10pt}
\end{table}

Besides, we evaluate the models $M_{\mathcal{L}_s}$ and $M_{\mathcal{L}_s+\mathcal{L}_{cs}}$ and the corresponding domain adapted models $M_{\mathcal{L}_s+\mathcal{L}_{GAN}}$ and $M_{\mathcal{L}_s+\mathcal{L}_{GAN}+\mathcal{L}_{cs}}$ on test set in source domain, and the results are shown in Table~\ref{tab:com_source}. As observed, the accuracies have little change before (96.40 and 96.75) and after (96.29 and 96.64) adaption, implying that our model can still keep good performance on source domain after domain adaption and simultaneously improve the performance on target domain. Therefore, the model does not need to know the source of images during testing and hence is more widely applicable.

\begin{figure*}[!thp]
    \centering
    \vspace{-5pt}
    \subfloat[$\mathcal{L}_{s}$ (68.9)]{
    \hspace{-25pt}
    \includegraphics[height=2.0in,width=2.0in]{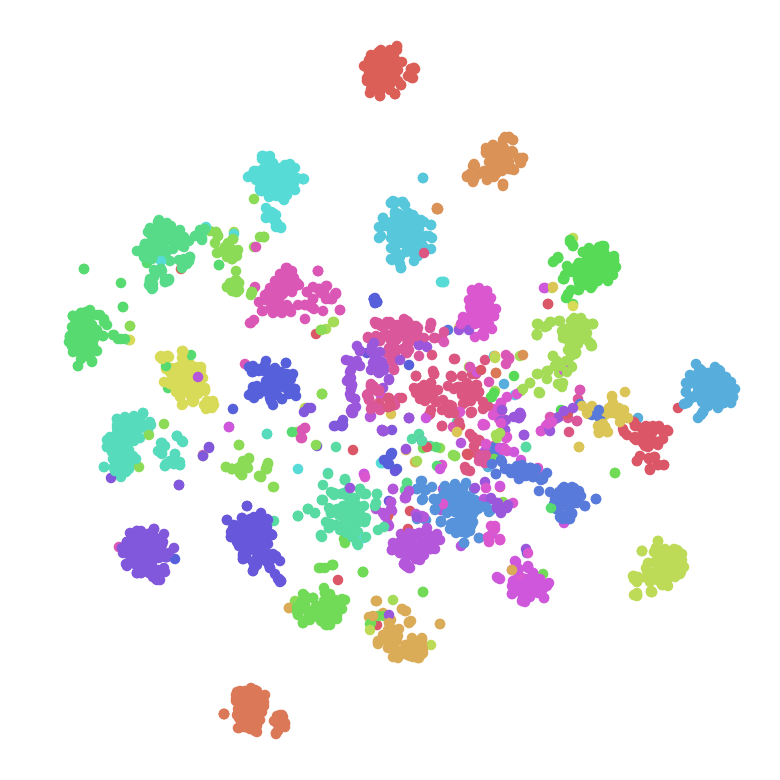}
    }
    \subfloat[$\mathcal{L}_{s}$+$\mathcal{L}_{GAN}$ (78.35)]{
    \hspace{-25pt}
    \includegraphics[height=2.0in,width=2.0in]{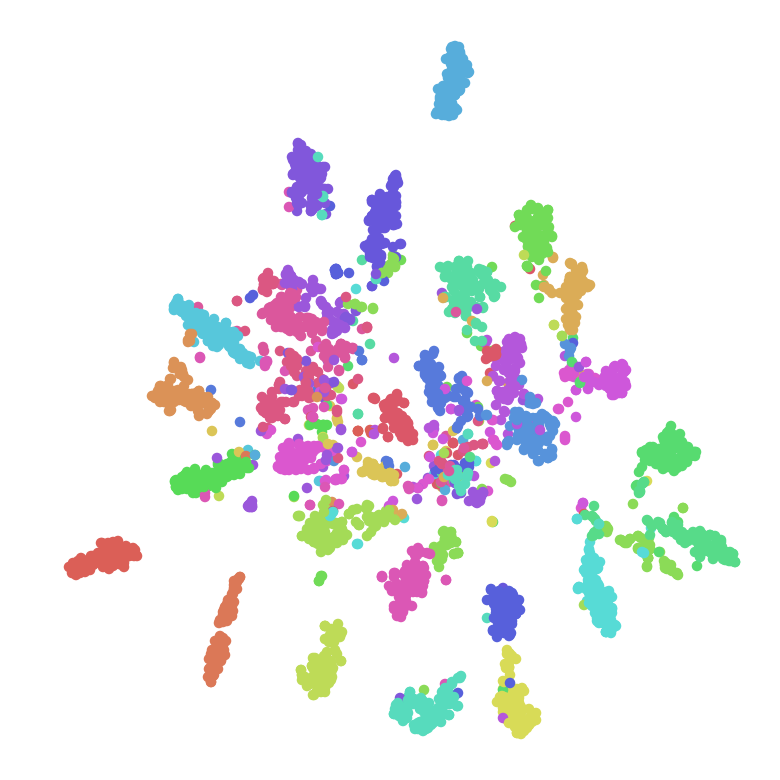}
    }
    \subfloat[$\mathcal{L}_{s}$+$\mathcal{L}_{GAN}$+$\mathcal{L}_{cs}$ (84.48)]{
    \hspace{-25pt}
    \includegraphics[height=2.0in,width=2.0in]{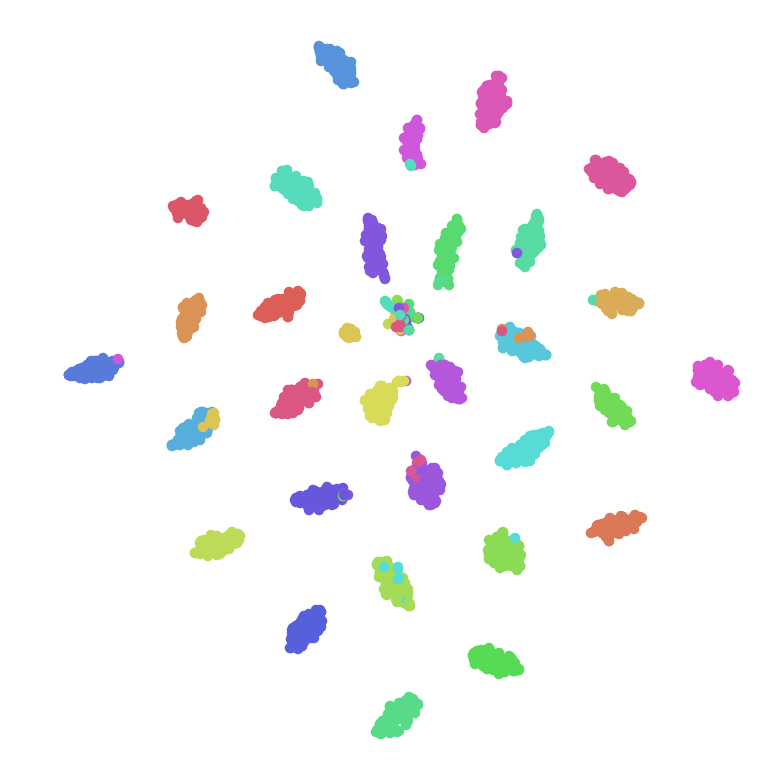}
    }
    \subfloat[$\mathcal{L}_{s}$+$\mathcal{L}_{GAN}$+$\mathcal{L}_{cs}$+$\mathcal{L}_{ct}$ (89.29)]{
    \hspace{-25pt}
    \includegraphics[height=2.0in,width=2.0in]{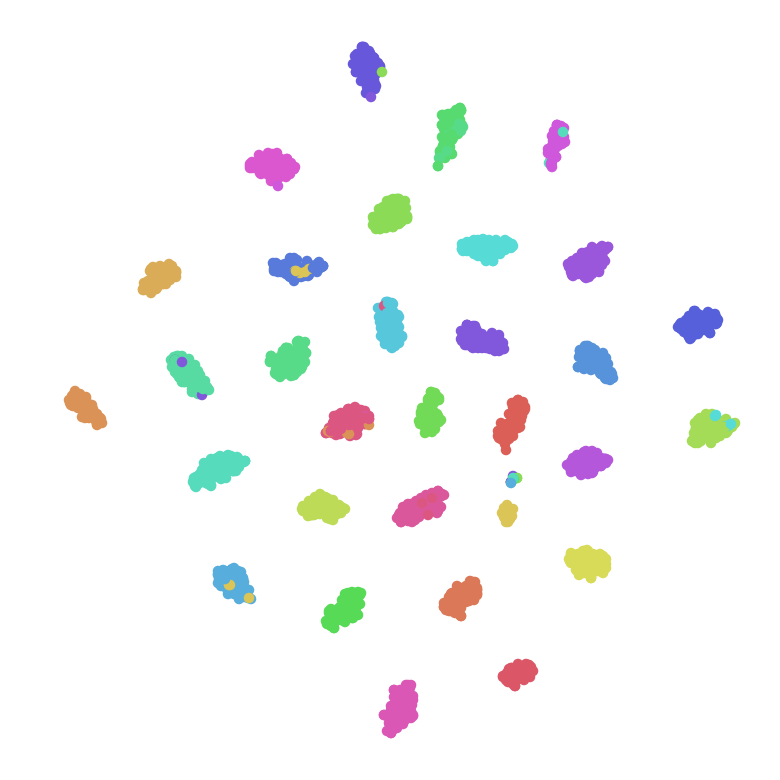}
    }\\
    \subfloat[$\mathcal{L}_{s}$ (68.9)]{
    \hspace{-25pt}
    \includegraphics[height=2.0in,width=2.0in]{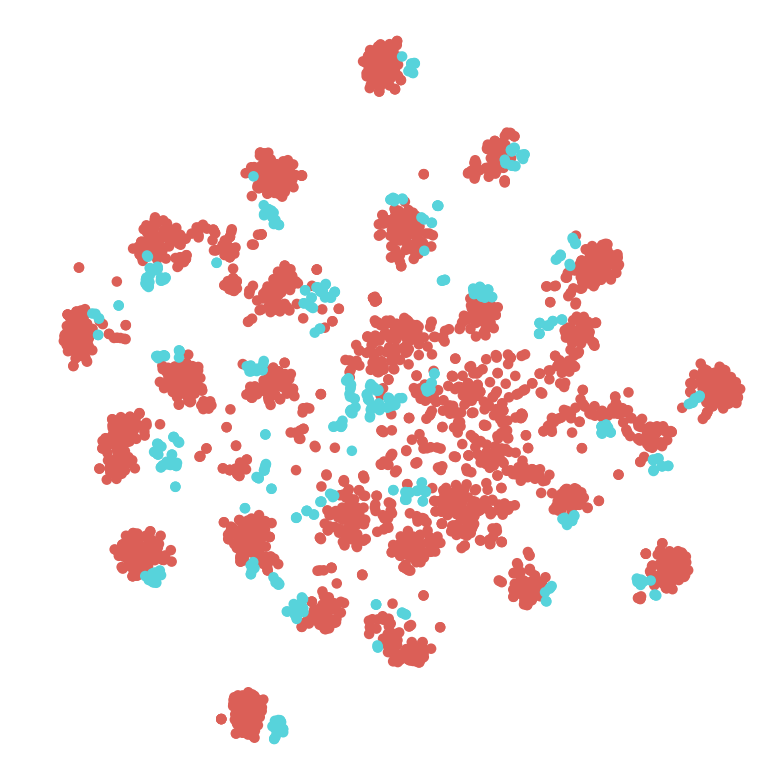}
    }
    \subfloat[$\mathcal{L}_{s}$+$\mathcal{L}_{GAN}$ (78.35)]{
    \hspace{-25pt}
    \includegraphics[height=2.0in,width=2.0in]{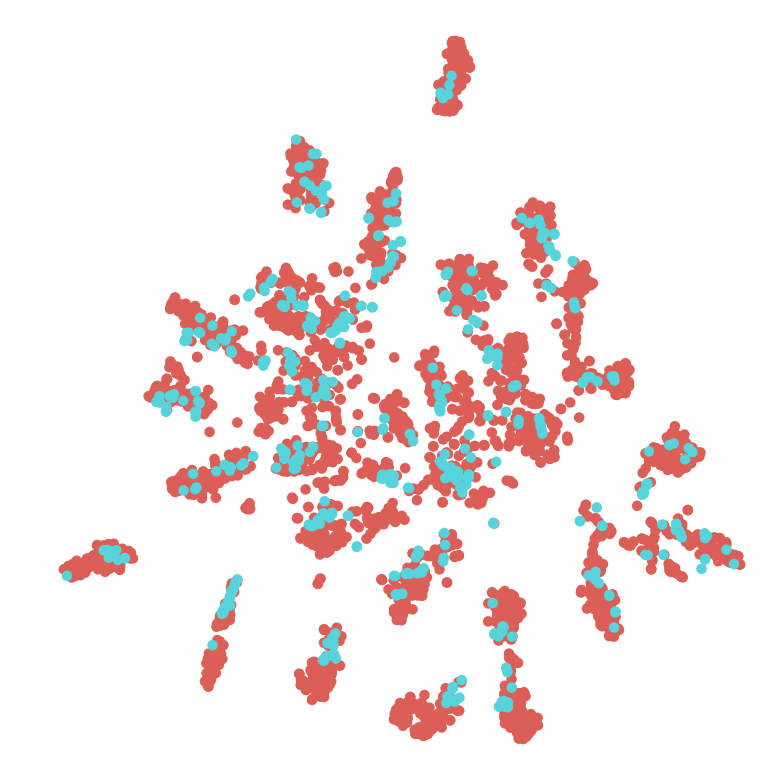}
    }
    \subfloat[$\mathcal{L}_{s}$+$\mathcal{L}_{GAN}$+$\mathcal{L}_{cs}$ (84.48)]{
    \hspace{-25pt}
    \includegraphics[height=2.0in,width=2.0in]{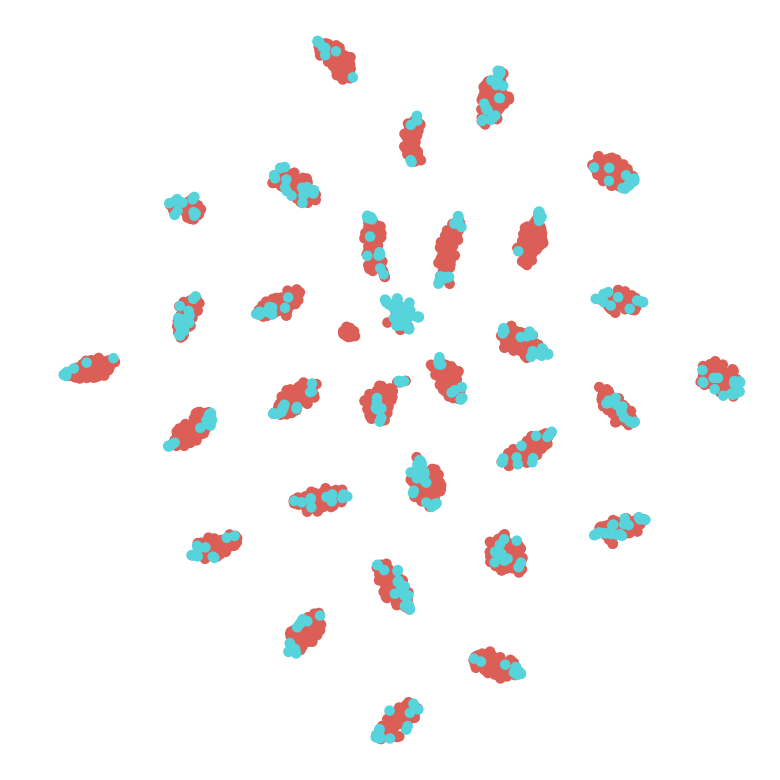}
    }
    \subfloat[$\mathcal{L}_{s}$+$\mathcal{L}_{GAN}$+$\mathcal{L}_{cs}$+$\mathcal{L}_{ct}$ (89.29)]{
    \hspace{-25pt}
    \includegraphics[height=2.0in,width=2.0in]{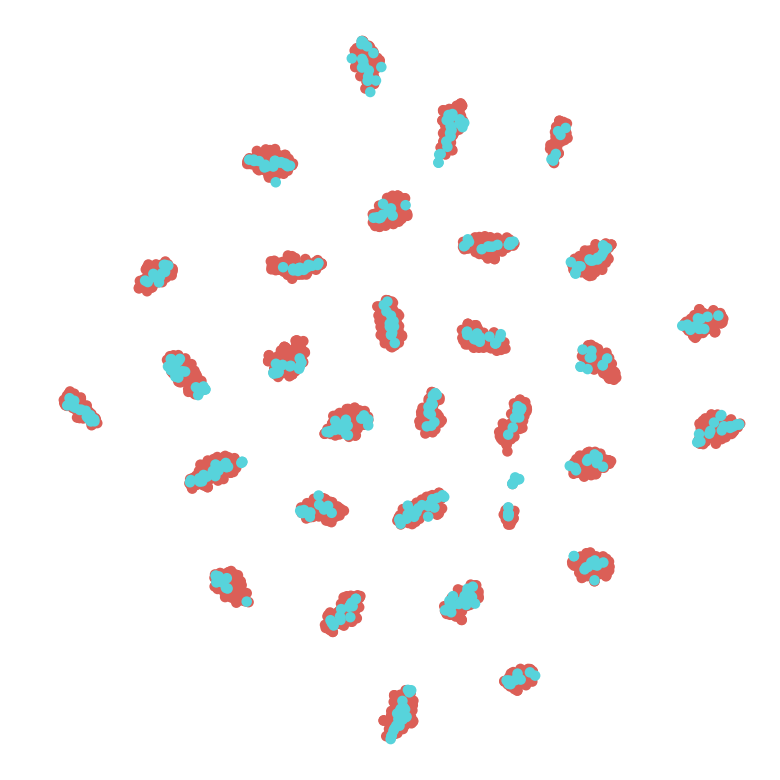}
    }
    \caption{The t-SNE visualization of features extracted by models trained with different loss functions for A$\rightarrow$D task (best viewed in color). In figures (a)-(d) we plot the distribution from category perspective and each color represents one class. In figures (e)-(h), we plot the distribution from domain perspective and points in red color represents samples from the source domain and blue points are samples from target domain. The values in parenthesis are corresponding accuracies.}
    \label{fig:visualize}
    \vspace{-10pt}
\end{figure*}

\subsubsection{Feature Visualization}
To understand the proposed DIAL network more intuitively, in Fig.~\ref{fig:visualize}, we visualize the t-SNE embeddings of features extracted by four models trained with different loss functions. In figures (a)-(d), we plot the distribution from category perspective and each color represents one category. In figures (e)-(h), we plot the distribution from domain perspective with red color and blue color representing source domain and target domain respectively. As shown in (a) and (e), the features of these two domains are separately distributed and there exists an obvious separation between features in two domains. In figures (b) and (f), after domain adaption, the features of two domains are merged in distribution and cannot be distinguished about the domain, which benefits from the domain-invariant feature extraction. However, there are many points scattered in the inter-class gap, whose labels may be misclassified with large possibility. After introducing the center loss on source samples, the features are more compacted and form clear clusters as observed in figures (c) and (g), indicating that the features learned are much more discriminative. Furthermore, since the model tries to extract domain-invariant features, the center loss for source samples also enforces the features in target domain like clusters. But they may fall into incorrect clusters, seeing the center cluster in figure (c), containing points of many colors. As shown in figures (d) and (h), after adding the $\mathcal{L}_{ct}$, the number of misclustered points decreased and more points fall into correct areas, which indicates that aligning the conditional distribution will guide more target samples to correct clusters. Through the feature visualization, we further validate the effectiveness of each component in the DIAL network.

\section{Conclusion}
In this paper, we propose a Domain-Invariant Adversarial  Learning (DIAL) network for unsupervised domain adaption, which is shown to extract both domain-invariant and discriminative features for source and target images. DIAL consists of an encoder, a classifier and an adversarial discriminator. The encoder is totally shared between the source and target domains, which is expected to learn domain-invariant features with the help of an adversarial discriminator. By sharing the encoder, the features of source and target domains can be jointly learned and the model does not need to know the source of images during testing and hence is more widely applicable. To learn discriminative features, we introduce the center loss. We also resort to the pesudo labels in target domain and align the conditional distributions of two domains. We evaluated the DIAL network on several unsupervised domain adaption benchmarks and achieved superior or comparable performance to state-of-the-art results.

{\small
\bibliographystyle{ieee}
\bibliography{egbib}
}

\end{document}